\documentclass[letterpaper]{article} 
\usepackage{aaai2026}  
\usepackage{times}  
\usepackage{helvet}  
\usepackage{courier}  
\usepackage[hyphens]{url}  
\usepackage{graphicx} 
\usepackage{natbib}  
\usepackage{caption} 
\usepackage{algorithm}
\usepackage{algorithmic}
\UseRawInputEncoding
\usepackage{newfloat}
\usepackage{amsmath,amssymb}
\usepackage{listings}
\DeclareCaptionStyle{ruled}{labelfont=normalfont,labelsep=colon,strut=off} 
\floatstyle{ruled}
\newfloat{listing}{tb}{lst}{}
\floatname{listing}{Listing}
\title{HyperLoad: A Cross-Modality Enhanced Large Language Model-Based Framework for Green Data Center Cooling Load Prediction}
\author{
    Haoyu Jiang\textsuperscript{\rm 1},
    Boan Qu\textsuperscript{\rm 2},
    Junjie Zhu\textsuperscript{\rm 1},
    Fanjie Zeng\textsuperscript{\rm 2},
    Xiaojie Lin\textsuperscript{\rm 1}\thanks{Corresponding author.},
    Wei Zhong\textsuperscript{\rm 1,3}
}
\affiliations{
    \textsuperscript{\rm 1}College of Energy Engineering, Zhejiang University, Hangzhou, China\\

    \textsuperscript{\rm 2}Polytechnic Institute, Zhejiang University, Hangzhou, China\\
    \textsuperscript{\rm 3}Shanghai Institute for Advanced Study, Zhejiang University, Shanghai, China

    \{haoyu.jiang, boan.qu, junjie\_zhu, zengfanjie, xiaojie.lin, wzhong\}@zju.edu.cn
}

\usepackage{bibentry}
\usepackage{multirow}
\usepackage{booktabs}
\usepackage{tabularx}
\usepackage{bibentry}

\begin{document}

\maketitle

\begin{abstract}

The rapid growth of artificial intelligence is exponentially escalating computational demand, inflating data center energy use and carbon emissions, and spurring rapid deployment of green data centers to relieve resource and environmental stress.  Achieving sub-minute orchestration of renewables, storage, and loads, while minimizing PUE and lifecycle carbon intensity, hinges on accurate load forecasting. However, existing methods struggle to address small-sample scenarios caused by cold start, load distortion, multi-source data fragmentation, and distribution shifts in green data centers. We introduce HyperLoad, a cross-modality framework that exploits pre-trained large language models (LLMs) to overcome data scarcity. In the Cross-Modality Knowledge Alignment phase, textual priors and time-series data are mapped to a common latent space, maximizing the utility of prior knowledge. In the Multi-Scale Feature Modeling phase, domain-aligned priors are injected through adaptive prefix-tuning, enabling rapid scenario adaptation, while an Enhanced Global Interaction Attention mechanism captures cross-device temporal dependencies. The public DCData dataset is released for benchmarking\footnote{\url{https://huggingface.co/datasets/Fine6868/DCData}}. Under both data sufficient and data scarce settings, HyperLoad consistently surpasses state-of-the-art (SOTA) baselines, demonstrating its practicality for sustainable green data center management.
\end{abstract}



\section{Introduction}

The rapid progress of artificial intelligence has caused computational demand to grow exponentially, doubling every 5.7 months since 2010, far exceeding hardware performance improvement \cite{Giattino2025}. Data from the Association for Computer Operations Management \cite{AFCOM2024} show that average rack power density in data centers increased from 6.1 kW in 2016 to 12 kW in 2024, imposing heavy stress on power and cooling systems. Traditional data centers, featuring “high-reliability, high-redundancy” designs and long-term high-load operation, further aggravate energy use and carbon emissions. By 2024, global data center electricity consumption reached 415 TWh, about 1.5\% of total global demand \cite{IEA2024}. Amid growing energy and environmental pressures, green data centers have become vital to reducing the ICT sector’s carbon footprint. Green data centers should serve as flexible load nodes that are deeply integrated with the power grid, enabling coordinated scheduling with photovoltaic, wind, and energy storage systems across time scales from seconds to minutes.  This interactive framework can sustain Power Usage Effectiveness below 1.3 and reduce lifecycle carbon emissions by over 30\%.

\begin{figure}[t]
\centering
\includegraphics[width=1.0\columnwidth]{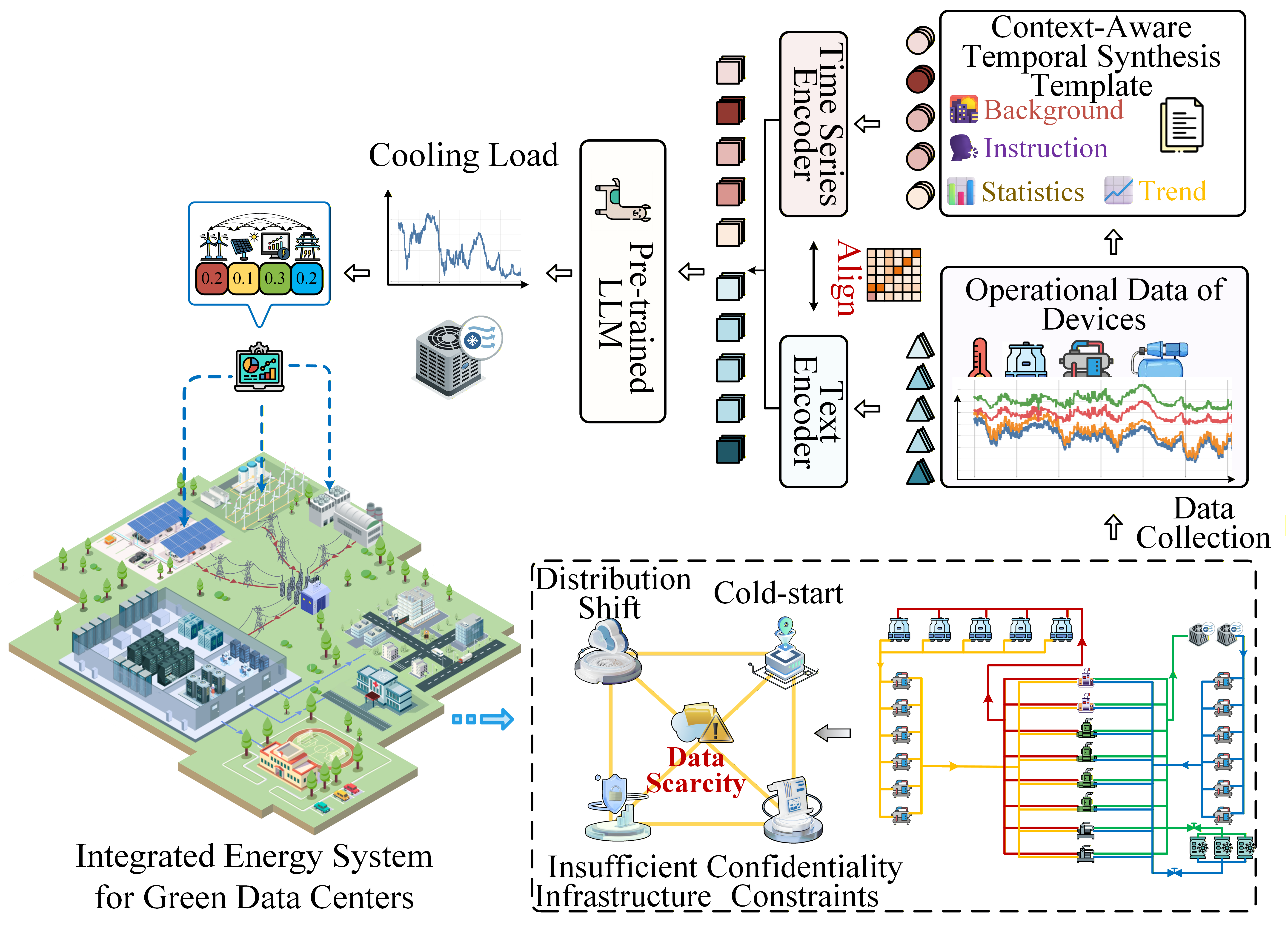} 
\caption{Schematic diagram of HyperLoad performing green data center cooling load prediction tasks.}
\label{dia}
\end{figure}

Field surveys reveal that conventional data centers depend on fixed-load control without dynamic regulation, making cooling systems unable to respond to real-time conditions, causing significant energy waste \cite{Radovanovic2022}. Cooling accounts for more than 25\% of total electricity use, the largest auxiliary load \cite{IEA2024}. Green data centers emphasize smart operation, where accurate cooling load forecasting is key to optimizing energy allocation, integrating renewables, and ensuring reliability. Precise forecasting aligns cooling load with variable renewable outputs, reducing curtailment, storage redundancy, and overall costs \cite{han2025_distributionally}. Yet emerging facilities face severe data scarcity. Limited operational history constrains model training, renewable integration alters data distributions, and multi-source data from logs and environmental monitoring are incomplete due to privacy and compliance barriers (Figure \ref{dia}).

\begin{figure}[t]
\centering
\includegraphics[width=1.0\columnwidth]{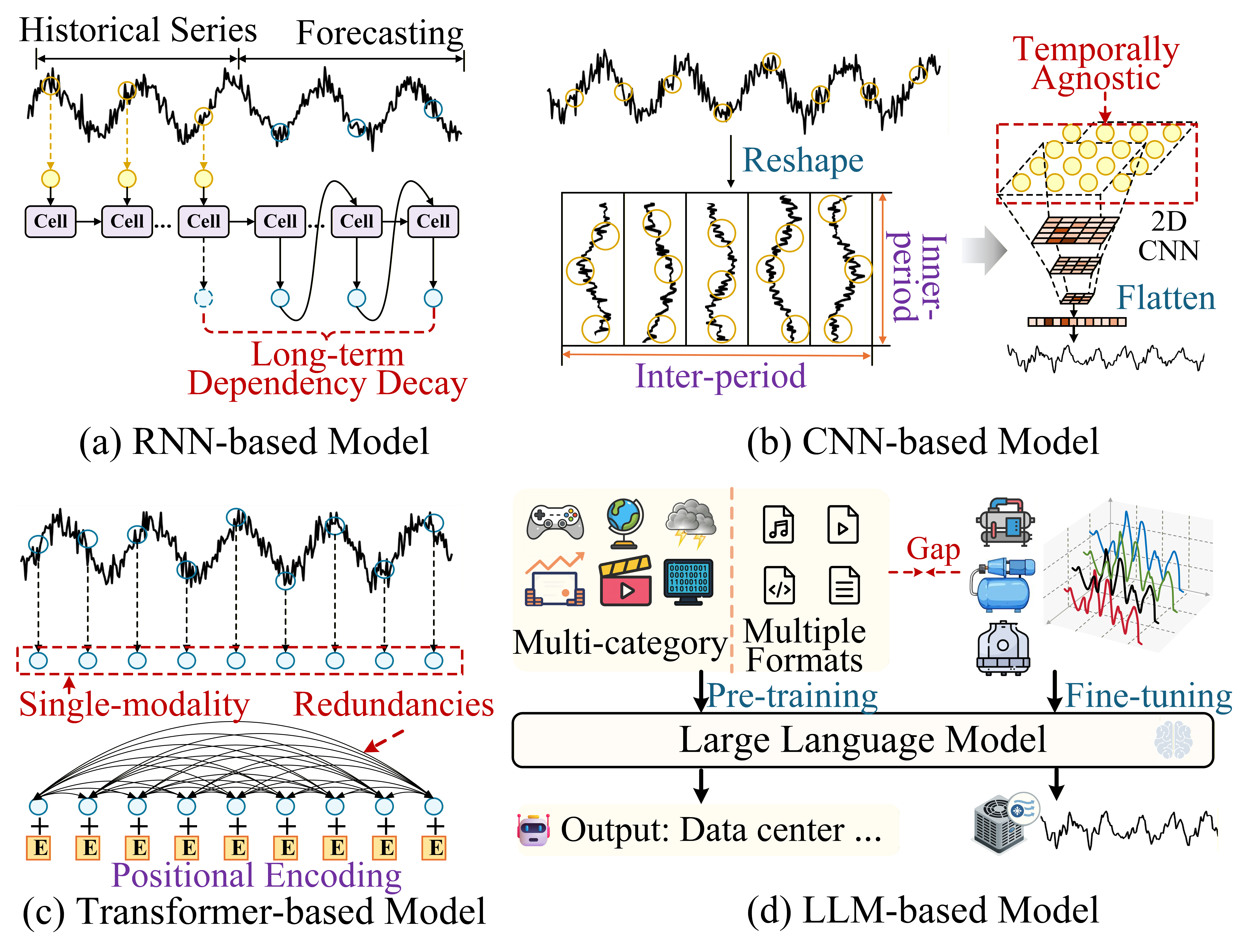} 
\caption{Limitations of time-series forecasting algorithms.}
\label{challenge}
\end{figure}

Current research on data center cooling load forecasting still relies heavily on engineers’ empirical rules, which depend on subjective judgment and lack quantitative rigor and replicability \cite{zhang2021_surveycooling}. Thermodynamic physical models can represent system mechanisms but are costly to deploy due to complex parameter measurement, calibration, and numerical computation \cite{lin2022_thermal}. With the rise of deep learning, models such as RNNs, CNNs, and Transformers have been introduced, markedly improving forecasting accuracy \cite{lu2022_loadforecasting} (Figure \ref{challenge}). RNN-based models (e.g., NGCU \cite{wang2022_ngcu}, SegRNN \cite{lin2023_segrnn}) capture short-term temporal relations but suffer from information decay over long sequences. CNN-based models (e.g., LightTS \cite{zhang2022_lightmlp}, TSMixer \cite{chen2023_tsmixer}) extract local features efficiently and support parallel computation but cannot preserve temporal order, weakening sequence dependency. Transformer-based models (e.g., Informer \cite{zhou2021_informer}) leverage attention mechanisms to learn temporal patterns, yet load variations depend on multiple heterogeneous factors, making single-modality modeling inadequate. Despite strong benchmark performance, these models require large, task-specific datasets for end-to-end training, and their learned representations are constrained by data distribution. When applied to green data centers with small samples, distribution shifts, and sparse labels, they often overfit and lose generalization, hindering fine-grained load prediction and intelligent operation.

Transformer-based LLMs such as GPT \cite{openai2023_gpt4} and LLaMA \cite{touvron2023_llama} are trained through self-supervised learning on massive text corpora, acquiring generalized representations that enable high-accuracy inference with minimal task-specific data. Building on their cross-domain success \cite{jiang2025_ucdr,bi2024_oceangpt}, recent research has explored applying LLMs to time-series forecasting by leveraging their sequence-pattern recognition and contextual-understanding abilities to address data scarcity. Models such as PromptCast \cite{xue2024_promptcast} and LFPLM \cite{gao2024_lfplm} focus on enhancing time-series feature representation, yet limited contextual reasoning reduces prediction accuracy. Time-LLM \cite{jin2024_timellm} and FPE-LLM \cite{qiu2024_fpellm} incorporate multi-modal fusion to guide LLM encoding, but directly processing raw text and temporal data ignores distributional and semantic disparities, weakening the value of textual priors. To better capture periodic variations, TEMPO  \cite{cao2024_tempo} applies STL decomposition to separate trend, seasonal, and residual components, though performance declines under complex seasonal dynamics \cite{bogahawatte2024_nncl}. Furthermore, green data centers exhibit nonlinear, dynamically evolving feature interactions among devices. Most existing methods fail to model these couplings or adaptively adjust feature weights, constraining their effectiveness in capturing intricate dependencies within complex operational datasets.

To address the foregoing challenges, we propose a cross-modality enhanced LLM-based load forecasting framework termed HyperLoad, which integrates cross-modality knowledge alignment with multi-scale feature modeling to enable efficient prediction in complex, dynamic settings (Figure \ref{dia}). The methodology unfolds in two phases:

1. \textit{Cross-Modality Knowledge Alignment phase}.

The Knowledge-Aligned Representation Integration (KARI) strategy is employed to dynamically adjust both the time series encoder and the text encoder, projecting text prior knowledge together with the load data features into a shared embedding space.  The alignment enhances cross-modalities consistency, enabling LLMs to better leverage textual knowledge for temporal reasoning.

2. \textit{Multi-Scale Feature Modeling phase}.

We propose an Adaptive Domain-specific Prefix Tuning (ADPT) strategy that leverages the text encoder’s cross-modal representational capacity acquired in the first phase to encode text prior knowledge and embed it as prefixes. Enables LLMs to efficiently adapt to new scenarios with limited data while capturing underlying load fluctuation mechanisms. To address device dependencies, we design an Enhanced Global Interaction Attention (EGIA) mechanism that further models cross-variable coupling, ensuring robustness under sample sparsity and distribution shifts. Finally, by integrating text priors with temporal features, the model harnesses the trend comprehension and reasoning capabilities learned from large-scale distributed data during pre-training to achieve accurate cooling load forecasting.

\begin{figure*}[t]
\centering
\includegraphics[width=0.89\textwidth]{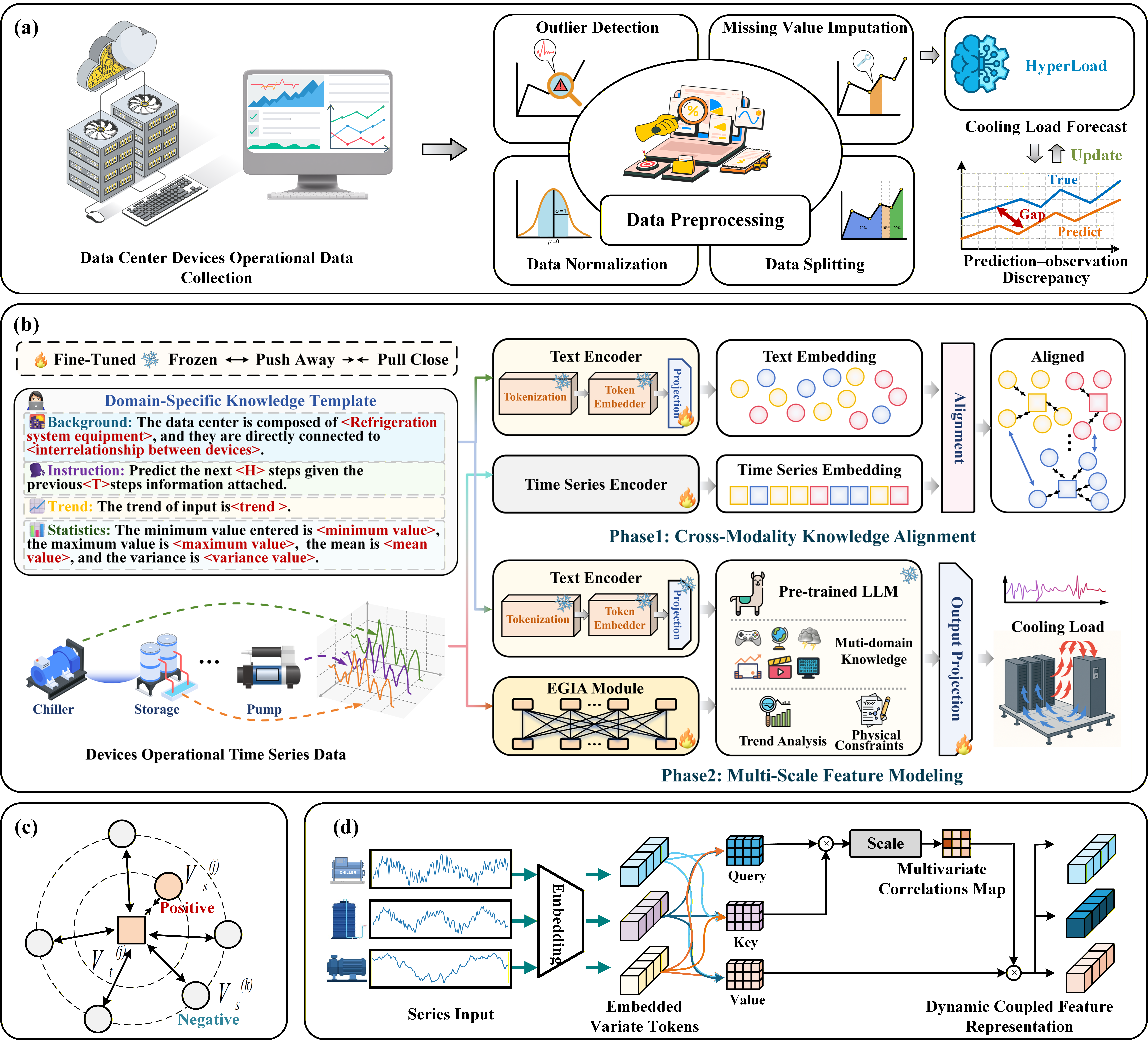} 
\caption{(a) Data collection and training process.
(b) Framework of HyperLoad.
(c) Schematic illustration of the KARI loss. (\fbox{\rule{0pt}{0.9ex}\rule{0.9ex}{0pt}} represents text modality, $\bigcirc$  represents time-series modality.)
(d) Schematic illustration of the EGIA mechanism.}
\label{hper_s}
\end{figure*}

The major contributions can be summarized as follows: 
\begin{itemize}

\item We construct the dataset for data center cooling load forecasting, DCData, comprising key parameters to provide a standardized, systematic foundation for model development and evaluation.

\item For the first time, we apply the trend-understanding and reasoning capabilities learned by LLMs from diverse distributions to the domain of green data center load forecasting, aiming to improve prediction accuracy and reduce the demand for large amounts of data.

\item We introduce a KARI strategy that reshapes the text encoder feature space to narrow distributional gaps between text priors and time-series data, enhancing prior knowledge utilization. Building on this, we propose an ADPT strategy that encodes critical green data center background as learnable prefix vectors, enabling rapid task adaptation without extensive retraining.

\item We propose an EGIA mechanism designed to efficiently capture the coupling relationships and global patterns across variables, thereby enhancing the model's ability to model key dependencies among devices.

\item Through experiments on DCData, HyperLoad outperforms SOTA baselines and demonstrates high accuracy and stability even under data scarcity.
\end{itemize}





\section{Methodology}
This section provides an overview of the proposed HyperLoad cooling load forecasting framework built on LLMs, as illustrated in Figure \ref{hper_s}(b). HyperLoad comprises two phases:

1. \textit{Cross-Modality Knowledge Alignment phase}: Using the KARI strategy to create a unified representation that merges textual prior knowledge with cooling load features.

2. \textit{Multi-Scale Feature Modeling phase}: Through an ADPT strategy, domain prior knowledge is injected into the LLMs, guiding the model to apply pre-training knowledge to green data center scenarios, thereby enabling precise encoding of load features. An EGIA mechanism further extracts multi-scale, fine-grained patterns and cross-variable dependencies from the time series. Ultimately, by harnessing the pre-trained LLMs’ capacities for trend analysis and logical reasoning, we achieve accurate cooling load forecasting.

\subsection{Problem Definition}

\noindent In the scenario of green data center cooling load forecasting, the historical observation data are denoted as  $\mathcal{X}\in \mathbb{R}^{{T}\times {M}}$, where $T$ represents the length of the sampling period and $M$ the number of monitored indicators. Given $L$ consecutive observations $x^{(j)}=(x_{1},x_{2},\dots,x_{L})\subset \mathcal{X}$, our objective is to predict the cooling load for the next $K$ time steps, denoted as $\hat{y}^{(j)}\in\mathbb{R}^{K\times 1}$.
The model parameters are optimized by minimizing the prediction error between $\hat{y}^{(j)}$  and the ground-truth values $y^{(j)}$.


\subsection{Cross-Modality Input Construction}
\paragraph{Construction of Time-Series Modality Inputs.}
Given the high dimensionality, multi-scale characteristics, and non-stationarity of data in green data center environments, coupled with substantial distributional variability across devices,  we apply reversible instance normalization to each feature column of $x^{(j)}$ to standardize them to zero mean and unit variance, thereby mitigating distributional heterogeneity. For the $i$-th feature column $x^{(j)}_{i}=\{x_{i,n}^{(j)}\}_{n=1}^{L}\!\in\!\mathbb{R}^{L\times 1}$, compute its mean $\mu_{i}^{(j)}$ and standard deviation $\sigma_{i}^{(j)}$:
 {\small
\begin{equation}
\mu_{i}^{(j)}=\frac{1}{L}\sum_{n=1}^{L}x_{i,n}^{(j)}, 
\qquad
\sigma_{i}^{(j)}=\sqrt{\frac{1}{L}\sum_{n=1}^{L}\bigl(x_{i,n}^{(j)}-\mu_{i}^{(j)}\bigr)^{2}} .
\label{eq:revin-stats}
\end{equation}}
For the $n$-th observation $x_{i,n}^{(j)}$ in this column, the normalized value is obtained as:
{\small
\begin{equation}
\bar{x}_{i,n}^{(j)}=\frac{x_{i,n}^{(j)}-\mu_{i}^{(j)}}{\sigma_{i}^{(j)}} .
\label{eq:revin-norm}
\end{equation}
}
The final normalized form of $x^{(j)}$ can thus be denoted as 
$\mathcal{P}^{(j)}=\bigl(\bar{x}^{(j)}_{1},\,\bar{x}^{(j)}_{2},\,\dots,\,\bar{x}^{(j)}_{M}\bigr).$

\paragraph{Construction of Text Modality Inputs.} 
Building on the template paradigm introduced in CLIP \cite{radford2021_clip}, we construct a \emph{Context-Aware Temporal Synthesis Template} for data center operations that serves as a priori embeddings for time-series data. By aligning text and time series representations within a shared latent space, we reinforce semantic coherence across the two modalities and steer the model toward capturing the intricate relationships among data center contexts, inter-device synergies, and load dynamics. 

First, we construct a domain knowledge base $\mathcal{T}^{(j)}_{b}$, which encompasses domain knowledge of data centers, including:


\textbf{A. Background}: Provide an overview of the data center system’s constituent modules together with their interdependencies, enabling the model to grasp the mechanisms of their coordinated operation.

\textbf{B. Instruction}: State the task specification unambiguously so that the model can correctly identify the prediction target and delineate the task boundaries.

Subsequently, we transform sequence $\mathcal{P}^{(j)}$ into its text counterpart $\mathcal{T}^{(j)}_{s}$, furnishing a structured input for later cross-modality alignment:

\textbf{C. Trend}: Describe the trend exhibited by the data segment, capturing long-term patterns and oscillations in cooling load to help model apprehend the dynamics of the series.

\textbf{D. Statistics}: Statistically analyze the data segment to obtain its extrema, mean, and variance, describing its distribution and fluctuations.

By concatenating the domain knowledge base $\mathcal{T}^{(j)}_{b}$ with the textual representation $\mathcal{T}^{(j)}_{s}$ of the time-series data, we obtain the \emph{Context-Aware Temporal Synthesis Template} ($\mathcal{T}^{(j)}_{CATS}$ ) for sequence $\mathcal{P}^{(j)}$:
\begin{equation}
    \mathcal{T}^{(j)}_{CATS} = \bigl[\,\mathcal{T}^{(j)}_{b},\; \mathcal{T}^{(j)}_{s}\bigr].
\end{equation}

\subsection{Cross-Modality Knowledge Alignment Phase}

During the Cross-Modality Knowledge Alignment phase, we introduce a KARI strategy that dynamically adjusts both the time series encoder and the text encoder. This strategy alleviates the distributional mismatch between the text and time-series modalities, thereby enabling more effective utilization of textual prior knowledge. After training, the text encoder is frozen and subsequently reused for all later training and inference phases.

\paragraph{Time-series modality feature encoding process.}

$\mathcal{P}^{(j)}$ is passed through the time series encoder ($\mathcal{F}_{\mathrm{TS}}(\cdot)$), which first embeds each variate into token-level representations (series representations of each token are extracted by the shared feed-forward network) and subsequently integrates them to derive the global time-series feature $\mathcal{V}_{s}^{(j)}$:
\begin{equation}
\mathcal{V}_{s}^{(j)} = \mathcal{F}_{\mathrm{TS}}\!\bigl(\mathcal{P}^{(j)}\bigr),\qquad \mathcal{V}_{s}^{(j)} \in \mathbb{R}^{d},
\end{equation}
where $d$ denotes the dimensionality of the features.

\paragraph{Text modality feature encoding process.}

The sequence $\mathcal{T}_{CATS}^{(j)}$ is processed by the text encoder to yield the text feature $\mathcal{V}_{t}^{(j)}$. The detailed procedure is as follows:

First, $\mathcal{T}_{CATS}^{(j)}$ is transformed into a discrete token sequence ($\mathcal{N}^{(j)}=(t_{1}^{(j)}, t_{2}^{(j)}, \dots, t_{n}^{(j)})$) through tokenization and token embedding.  On this basis, a feed-forward network ($\mathcal{E}$) maps each token $t_{i}^{(j)}$ to a learnable embedding $e_{i}^{(j)}$, thereby yielding the embedding sequence:
\begin{equation}
\mathcal{D}^{(j)}= \mathcal{E}\bigl(\mathcal{N}^{(j)}\bigr)=(e_{1}^{(j)}, e_{2}^{(j)}, \dots, e_{n}^{(j)}),
\end{equation}
a global representation $\mathcal{V}_t^{(j)} \in \mathbb{R}^{d}$ is then extracted from the encoded sequence for downstream cross-modal alignment.





\paragraph{KARI Strategy.}

In the field of green data center load forecasting, the text modality (including load feature descriptions, domain knowledge bases, and task specifications) and the time-series modality (expressed as dynamic numerical sequences) differ fundamentally in both data distribution and semantic space. The former encodes structured knowledge through discrete symbols, whereas the latter represents complex system dynamics via continuous signals. This modal heterogeneity leads to a representational mismatch during feature-level fusion, thereby weakening the guiding role of textual prior knowledge in modeling load data. 


To address the above challenge, this study proposes a KARI loss ($\mathcal{L}_{\mathcal{KARI}}$) that projects the feature representations of the time-series $\mathcal{V}_s^{(j)}$ and text $\mathcal{V}_t^{(j)}$ modalities into a unified shared latent space (see Figure \ref{hper_s}(c)). In doing so, it ensures that the features produced by the text encoder become distributionally compatible with the time-series load data, allowing text priors to supply effective inductive support to the load encoding process and enhancing the model’s understanding of the mechanisms driving load fluctuations. $\mathcal{L}_{\mathrm{KARI}}$ is defined as follows:
{\footnotesize
\begin{equation}
\mathcal{L}_{\mathcal{KARI}}
= -\,\log
\frac{\exp\bigl(\mathcal{C}\!\bigl(\mathcal{V}_t^{(j)},\,\mathcal{V}_s^{(j)}\bigr)/\tau\bigr)}
{\sum_{k=1}^{\mathcal{B}}\exp\bigl(\mathcal{C}\!\bigl(\mathcal{V}_t^{(j)},\,\mathcal{V}_s^{(k)}\bigr)/\tau\bigr)},
\end{equation}
}
where \(\mathcal{C}(\,\cdot\,,\,\cdot\,)\) denotes cosine similarity and \(\tau\) the temperature (fixed to 0.05), $\mathcal{B}$ denotes batch size. The \(\mathcal{L}_{\mathcal{KARI}}\) minimizes the distance between each sequence feature and its corresponding text feature, thereby projecting them into a shared feature space and achieving cross-modality semantic alignment with efficient feature representations.


\subsection{Multi-Scale Feature Modeling Phase}
During the multi-scale feature modeling stage, we introduce an ADPT strategy and an EGIA mechanism. ADPT strategy injects domain prior knowledge into the LLMs in an efficient manner, guiding it to fully leverage knowledge accumulated during pre-training so as to encode load features with higher precision; EGIA mechanism strengthens the model’s capacity to capture multi-scale fine-grained patterns and complex cross-variable dependencies. Ultimately, by exploiting the LLMs’ strengths in trend analysis and reasoning, the framework achieves efficient prediction of cooling load.

\paragraph{ADPT Strategy.} We first use the text encoder trained in the Cross-Modality Knowledge Alignment phase to encode the \textit{Context-Aware Temporal Synthesis Template} adaptively constructed for each data patch. The resulting template encoding $\mathcal{V}_{T}^{(j)}$ is then used as a prefix and concatenated with the corresponding time-series encoding, and combined sequence is fed into LLMs. The objectives of this strategy are: 
\begin{itemize}
\item \textbf{Knowledge transfer:} Guide the LLMs to apply the trend-understanding and reasoning capabilities accumulated during pre-training to problems in the data center.
\item \textbf{Contextual understanding:} Enhance the LLMs’ grasp of the intrinsic relationships between the coordinated operation of devices and fluctuations in cooling load.
\end{itemize}

\paragraph{EGIA Mechanism.} 
In green data center scenarios, multiple environmental variables (e.g., inlet and outlet cooling-water temperature, air humidity, and equipment load) exhibit pronounced, temporally evolving coupling. Existing approaches primarily emphasize modeling temporal dependencies while overlooking inter-device collaboration and the adaptive allocation of informational weights, thereby limiting their effectiveness for load data with nonlinear and dynamically evolving characteristics. To address this issue, we design an EGIA mechanism that jointly learns multivariate sequence representations and their adaptive correlations, thereby elucidating the impact of device collaboration on cooling load fluctuations. The EGIA mechanism is illustrated in Figure \ref{hper_s}(d).
Specifically, we first embed the time series of each device-level variable in $\mathcal{P}^{(j)}$ as an independent token to obtain the feature representation $\mathcal{S}^{(j)}$:
\begin{equation}
\mathcal{S}^{(j)} = \mathrm{Embedding}\!\bigl(\mathcal{P}^{(j)}\bigr), 
\qquad \mathcal{S}^{(j)} \in \mathbb{R}^{M \times d}.
\end{equation}

The embedded features are linearly projected into query ($\mathcal{Q}^{(j)}$), key ($\mathcal{K}^{(j)}$), and value ($\mathcal{H}^{(j)}$) representations via three learnable projection matrices $\mathcal{W}_{Q},\mathcal{W}_{K},\mathcal{W}_{H}\in\mathbb{R}^{d\times d}$:

{\small
\begin{equation}
\mathcal{Q}^{(j)} = \mathcal{S}^{(j)}\mathcal{W}_{Q}, 
\mathcal{K}^{(j)} = \mathcal{S}^{(j)}\mathcal{W}_{K}, 
\mathcal{H}^{(j)} = \mathcal{S}^{(j)}\mathcal{W}_{H},
\end{equation}}

Subsequently, we model the correlations among variables, such that highly correlated variables receive greater weights in their subsequent interaction with the value representation:
{\fontsize{7pt}{9pt}\selectfont\begin{equation}
\mathcal{V}_{E}^{(j)} 
= \mathcal{A}ttn\!\bigl(\mathcal{Q}^{(j)},\mathcal{K}^{(j)},\mathcal{H}^{(j)}\bigr)
= \mathrm{Softmax}\!\Bigl(\frac{\mathcal{Q}^{(j)}\mathcal{K}^{(j)\!\top}}{\sqrt{d}}\Bigr)\mathcal{H}^{(j)}.
\end{equation}}


\paragraph{Pre-trained LLM-based Backbone.} We adopt LLaMA-7B as the backbone model of HyperLoad. To maintain its pre-trained capacity for trend inference and reasoning, all parameters aside from the output prediction head are frozen while training. Fine-tuning procedure proceeds as follows:

First, we apply the ADPT strategy to concatenate the multimodal features \(\mathcal{V}_{T}^{(j)}\) and \(\mathcal{V}_{E}^{(j)}\), constructing the input:
\begin{equation}
\mathcal{V}_{\text{in}}^{(j)} = [\mathcal{V}_{T}^{(j)},\, \mathcal{V}_{E}^{(j)}].
\end{equation}
Subsequently, \(\mathcal{V}_{\text{in}}^{(j)}\) is fed into the pre-trained LLM \(\bigl(\mathcal{O}_{\theta}^{\mathrm{LLM}}\bigr)\) to obtain the encoded representation, and the output is obtained through a trainable linear projection layer ($\mathcal{F}_{proj}()$):
\begin{equation}
\mathcal{V}_{\text{out}}^{(j)} = \mathcal{F}_{proj}\bigl(\mathcal{O}_{\theta}^{\mathrm{LLM}}\!\bigl(\mathcal{V}_{\text{in}}^{(j)}\bigr)\bigr).
\end{equation}

Finally, the output is inverse-normalized to obtain the final prediction \(\hat{y}^{(j)}\). We adopt the MSE as the training objective to quantify the discrepancy between the predicted values \(\hat{y}^{(j)}\) and the ground-truth values \(y^{(j)}\), and perform gradient updates accordingly:
{
\small
\begin{equation}
\mathcal{L}_{MSE} = \frac{1}{K} \sum_{k=1}^{K} \left( \hat{y}^{(j)}_{k} - y^{(j)}_{k} \right)^2.
\end{equation}
}

\begin{table*}[t]
\centering
{\fontsize{9pt}{10pt}\selectfont 
 \resizebox{1.0\textwidth}{!}{
\begin{tabular*}{\textwidth}{@{\extracolsep{\fill}} c|l|cc|cc|cc|cc}
\toprule
Training Data &  \multirow{2}{*}{Methods} &
\multicolumn{2}{c|}{PL = 12} &
\multicolumn{2}{c|}{PL = 24} &
\multicolumn{2}{c|}{PL = 48} &
\multicolumn{2}{c}{PL = 96} \\
\cmidrule(lr){3-10}
(Percent) & & MSE & MAE & MSE & MAE & MSE & MAE & MSE & MAE \\
\midrule
\multirow{11}{*}{100\%} & Autoformer \cite{wu2021_autoformer}  & 0.0384 & 0.1307 & 0.0518 & 0.1613 & 0.1383 & 0.2613 & 0.0656 & 0.1944 \\
 & FreTS    \cite{yi2023_freqmlp}    & 0.0065 & 0.0572 & 0.0093 & 0.0678 & 0.0166 & 0.0942 & 0.0270 & 0.1270 \\
 & iTransformer \cite{liu2024_itransformer}   & 0.0065 & 0.0589 & 0.0108 & 0.0722 & 0.0153 & 0.0854 & 0.0203 & 0.0977 \\
 & LightTS  \cite{zhang2022_lightmlp}    & 0.0070 & 0.0618 & 0.0103 & 0.0738 & 0.0240 & 0.1196 & 0.0410 & 0.1431 \\
 & N\_Transformer \cite{liu2022_nonstationary}& 0.0096 & 0.0703 & 0.0130 & 0.0978 & 0.0144 & 0.0865 & 0.0240 & 0.1109 \\
 & PatchTST  \cite{nie2023_ts64words}   & 0.0075 & 0.0652 & 0.0105 & 0.0686 & 0.0145 & 0.0839 & 0.0293 & 0.1005 \\
 & PAttn   \cite{tan2024arelmtsf}     & 0.0061 & 0.0569 & 0.0094 & 0.0692 & 0.0158 & 0.0883 & 0.0218 & 0.1008 \\
 & SCINet    \cite{liu2022_scinet}   & 0.0063 & 0.0579 & 0.0088 & 0.0661 & 0.0311 & 0.0783 & 0.0271 & 0.1025 \\
 & Transformer \cite{vaswani2017_attention} & 0.0872 & 0.1816 & 0.1563 & 0.3460 & 0.1043 & 0.2781 & 0.1496 & 0.2292 \\
 & TSMixer  \cite{chen2023_tsmixer}    & 0.0134 & 0.0870 & 0.0229 & 0.1301 & 0.0510 & 0.1793 & 0.2822 & 0.4680 \\
 & TimesNet   \cite{wu2023_timesnet}  & 0.0062 & 0.0565 & 0.0098 & 0.0708 & 0.0155 & 0.0879 & 0.0308 & 0.1262 \\
 & \textbf{HyperLoad}    & \textbf{0.0055} & \textbf{0.0539} & \textbf{0.0084} & \textbf{0.0637} & \textbf{0.0124} & \textbf{0.0748} & \textbf{0.0192} & \textbf{0.0931} \\
\midrule
\multirow{11}{*}{50\% } & Autoformer \cite{wu2021_autoformer}  & 0.0315 & 0.1370 & 0.0381 & 0.1470 & 0.0434 & 0.1685 & 0.0691 & 0.2070 \\
 & FreTS    \cite{yi2023_freqmlp}     & 0.0071 & 0.0595 & 0.0096 & 0.0683 & 0.0134 & 0.0796 & 0.0208 & 0.0997 \\
 & iTransformer \cite{liu2024_itransformer}  & 0.0067 & 0.0592 & 0.0118 & 0.0761 & 0.0139 & 0.0809 & 0.0211 & 0.0991 \\
 & LightTS    \cite{zhang2022_lightmlp}  & 0.0077 & 0.0659 & 0.0181 & 0.1074 & 0.0280 & 0.1334 & 0.0510 & 0.1878 \\
 & N\_Transformer \cite{liu2022_nonstationary} & 0.0075 & 0.0624 & 0.0101 & 0.0705 & 0.0154 & 0.0866 & 0.0223 & 0.1050 \\
 & PatchTST  \cite{nie2023_ts64words}   & 0.0062 & 0.0565 & 0.0090 & 0.0669 & 0.0149 & 0.0836 & 0.0227 & 0.1046 \\
 & PAttn    \cite{tan2024arelmtsf}     & 0.0068 & 0.0599 & 0.0093 & 0.0682 & 0.0135 & 0.0812 & 0.0220 & 0.1020 \\
 & SCINet  \cite{liu2022_scinet}     & 0.0067 & 0.0593 & 0.0096 & 0.0691 & 0.0139 & 0.0807 & 0.0217 & 0.1020 \\
 & Transformer \cite{vaswani2017_attention} & 0.1323 & 0.3184 & 0.5614 & 0.7100 & 1.0982 & 0.9328 & 1.8995 & 1.2011 \\
 & TSMixer   \cite{chen2023_tsmixer}    & 0.0402 & 0.1564 & 0.0991 & 0.2255 & 0.3436 & 0.4792 & 0.6154 & 0.6494 \\
 & \textbf{HyperLoad}    & \textbf{0.0058} & \textbf{0.0544} & \textbf{0.0087} & \textbf{0.0638} & \textbf{0.0128} & \textbf{0.0757} & \textbf{0.0200} & \textbf{0.0950} \\
\midrule
\multirow{11}{*}{25\% }& Autoformer \cite{wu2021_autoformer} & 0.1171 & 0.2413 & 0.1368 & 0.2547 & 0.1519 & 0.2944 & 0.1782 & 0.3519 \\
 & FreTS \cite{yi2023_freqmlp}  & 0.0238 & 0.1203 & 0.0351 & 0.1469 & 0.0454 & 0.1617 & 0.0760 & 0.2029 \\
 & iTransformer \cite{liu2024_itransformer}  & 0.0177 & 0.0978 & 0.0227 & 0.1082 & 0.0299 & 0.1227 & 0.0454 & 0.1499 \\
 & LightTS \cite{zhang2022_lightmlp} & 0.0202 & 0.1087 & 0.0514 & 0.1867 & 0.1025 & 0.2702 & 0.2133 & 0.3826 \\
 & N\_Transformer \cite{liu2022_nonstationary} & 0.0168 & 0.0927 & 0.0269 & 0.1153 & 0.1078 & 0.2592 & 0.0937 & 0.2185 \\
 & PatchTST \cite{nie2023_ts64words} & 0.0241 & 0.1011 & 0.0325 & 0.1305 & 0.0305 & 0.1220 & 0.0540 & 0.1663 \\
 & PAttn \cite{tan2024arelmtsf}  & 0.0203 & 0.1051 & 0.0268 & 0.1179 & 0.0360 & 0.1358 & 0.0499 & 0.1583 \\
 & SCINet \cite{liu2022_scinet} & 0.0302 & 0.1269 & 0.0300 & 0.1306 & 0.0424 & 0.1470 & 0.0561 & 0.1700 \\
 & Transformer \cite{vaswani2017_attention} & 2.4365 & 1.4666 & 4.1371 & 1.9232 & 4.4457 & 1.9520 & 6.006 & 2.3439 \\
 & TSMixer \cite{chen2023_tsmixer}  & 0.5039 & 0.6698 & 0.8543 & 0.8881 & 3.4351 & 1.6968 & 5.8421 & 2.0656 \\
 & \textbf{HyperLoad} & \textbf{0.0140} & \textbf{0.0839} & \textbf{0.0197} & \textbf{0.0969} & \textbf{0.0294} & \textbf{0.1170} & \textbf{0.0442} & \textbf{0.1435} \\
\bottomrule

\end{tabular*}
}
\caption{Prediction performance using 100\%, 50\%, and 25\% of the training data.}
\label{tab:results}}
\end{table*}

\section{Experiments}
\subsection{Experimental Setup}

High-quality public datasets are scarce because of privacy rules and limited infrastructure. To bridge this gap and evaluate HyperLoad, we collected 5-minute interval data from a data center in Dongguan spanning October to December, 2024. The dataset comprises 41 variables, including: outdoor temperature and humidity (2 groups); inlet/outlet temperatures of cooling water (9 cooling pumps); inlet/outlet temperatures of chilled water (9 chiller pumps); cooling load. After removing segments affected by commissioning and equipment faults, and applying a series of preprocessing steps (see Figure \ref{hper_s}(a)), we obtained the DCData dataset (contains 13,438 records). It is chronologically partitioned training, validation, and test sets in a 7:1:2 ratio. We assess HyperLoad in two data-availability settings: 
\begin{itemize}
\item \textbf{Data sufficient:} Model is trained on full training set to characterize the upper bound of achievable performance.
\item \textbf{Data scarce:} Model is trained using 50\% and 25\% of training set (partitioned in chronological order), to simulate limited-data conditions and assess its potential applicability in green data centers.
\end{itemize}

Implemented in PyTorch and trained on a single NVIDIA A100 GPU, the model employed an input sequence length of 96 and  forecasting length of 12, 24, 48, and 96 steps. Key hyper-parameters included an initial learning rate of \( 7 \times 10^{-4} \), batch size $\mathcal{B}$ = 64, and an 8 layer LLaMA backbone, with optimization via Adam. 


\subsection{Experimental Results}

\subsubsection{Model Performance in the Data Sufficient Setting.} First, we evaluate all models with input length of 96 and prediction lengths (PL) of 12, 24, 48, and 96. The results are summarized in top of Table \ref{tab:results}, with the best scores highlighted in bold. HyperLoad consistently outperforms all SOTA baselines across all forecast lengths, underscoring its advantage for cooling load forecasting. Specifically, at the 96-step prediction length, HyperLoad reduces MAE by 52.1\%, 26.7\%, 4.7\%, 34.9\%, 16.1\%, 7.4\%, 7.6\%, 9.2\%, 59.4\%, and 26.2\% relative to Autoformer, FreTS, iTransformer, LightTS, N\_Transformer, PatchTST, PAttn, SCINet, Transformer, and TimesNet, respectively. Corresponding MSE reductions are 70.7\%, 28.9\%, 5.4\%, 53.2\%, 20.0\%, 34.5\%, 11.9\%, 29.1\%, 87.2\%, and 37.7\%, respectively. 



\begin{figure}[t]
\centering
\includegraphics[width=1.0\columnwidth]{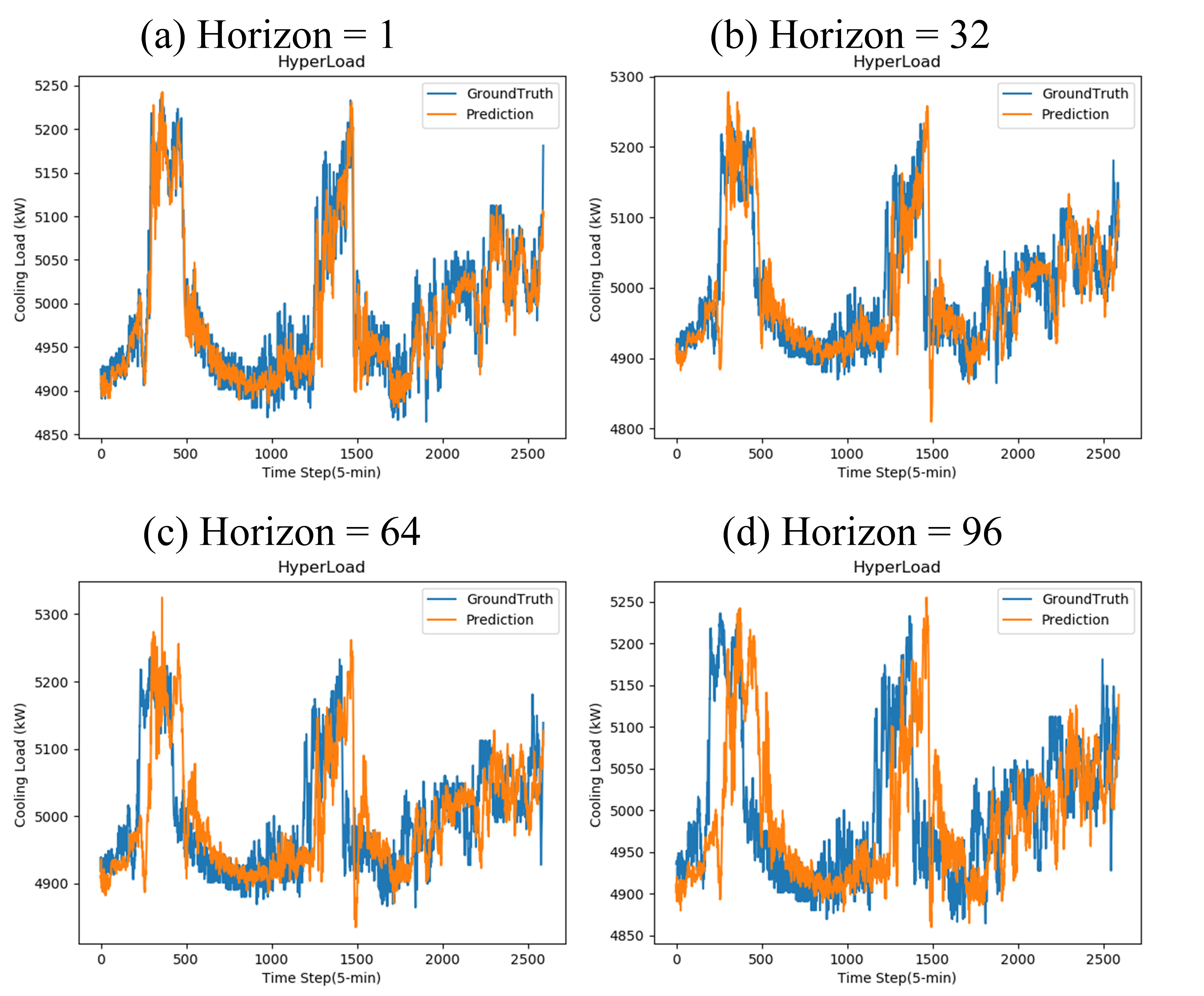} 
\caption{Prediction results for different horizons.}
\label{result}
\end{figure}

In experiments with both input and prediction sequences fixed at 96 time steps, HyperLoad was evaluated over horizons of 1, 32, 64, and 96 (S-Table 1). It achieved the highest accuracy at all horizons. Visualizations (Figure \ref{result}, S-Figures 1–4) show that HyperLoad consistently follows target trends more closely than other models, confirming its superior and stable forecasting performance across different time steps, demonstrates its ability to accurately perform predictions at varying horizons according to practical requirements.
\subsubsection{Model Performance in the Data Scarce Setting.}

Owing to the challenges imposed by data scarcity, the performance of all models declined. Nevertheless, HyperLoad capitalizes on the prior knowledge accumulated during pre-training to make accurate predictions of cold-start data center load demand with only a limited quantity of data, and its performance is correspondingly less affected. Concretely, across the four forecasting lengths examined, HyperLoad outperforms SOTA baselines in the data scarce setting where 50\% of the training data are provided (middle of Table \ref{tab:results}), reducing the MSE by 6.5\%, 3.3\%, 4.5 \%, and 3.9\% and the MAE by 3.7\%, 4.6\%, 4.9\%, and 4.1\%, respectively. Under an extreme data scarce setting with only 25\% of the training data (bottom of Table \ref{tab:results}), the corresponding reductions in MSE are 16.7\%, 13.2\%, 1.7\%, and 2.6\%, while MAE decreases by 9.5\%, 10.5\%, 4.1\%, and 4.3\%, respectively. 

\begin{figure}[t]
\centering
\includegraphics[width=1.0\columnwidth]{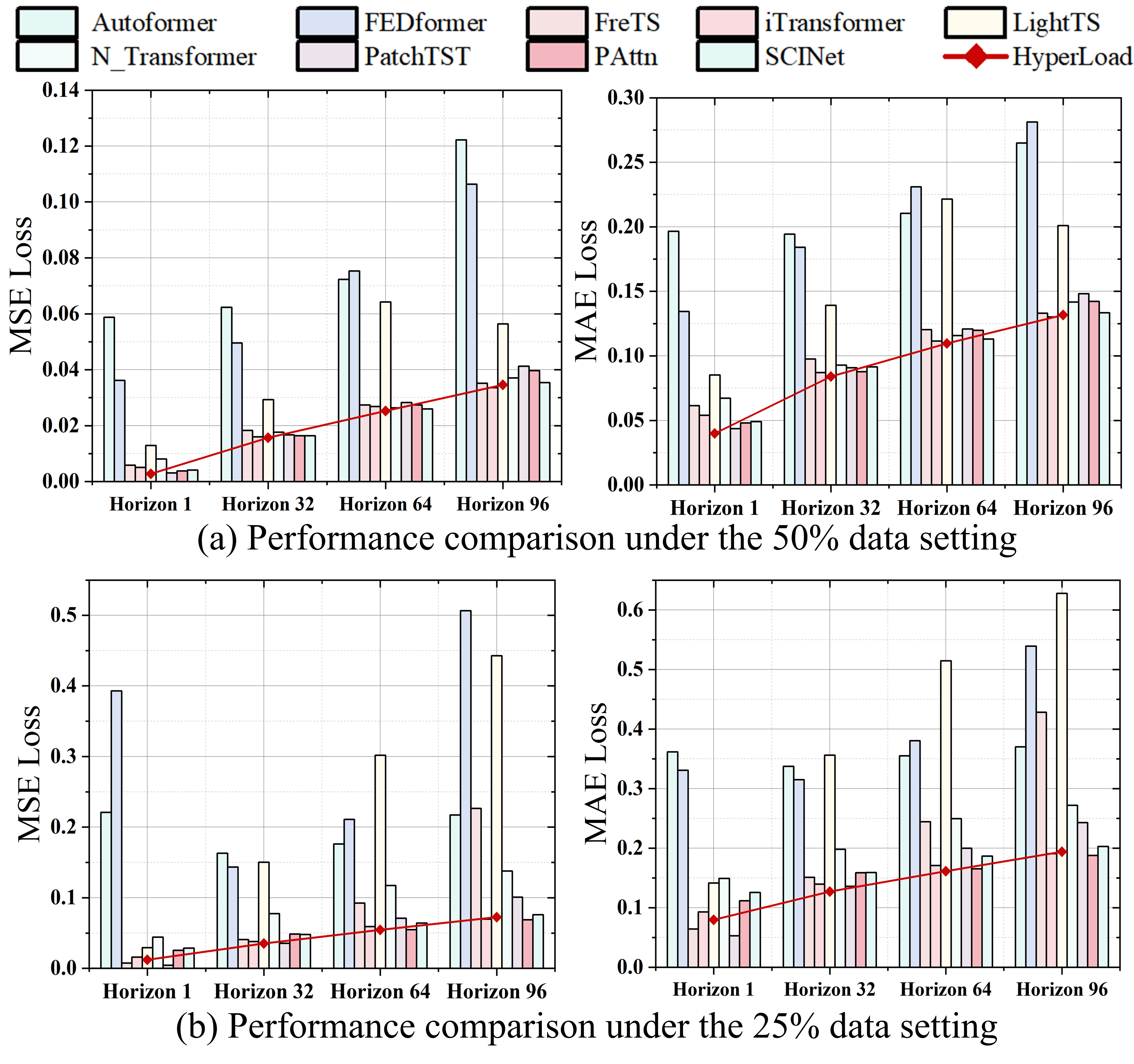} 
\caption{Model performance in the data scarce setting.}
\label{result_sca}
\end{figure}

Additionally, we evaluated each method with both the input and prediction sequence lengths fixed at 96 across four forecasting horizons (1, 32, 64, 96). The corresponding results are shown in Figure \ref{result_sca}. Under all settings, HyperLoad demonstrates consistently strong predictive performance and achieves the best overall results. This shows that HyperLoad effectively transfers LLMs’ trend-understanding and reasoning capabilities to data center load forecasting, mitigating data scarcity in real-world settings.


\subsubsection{Ablation Study.} To validate the effectiveness of the HyperLoad framework, we conducted ablation experiments on the full training dataset using an input  length of 96 and an output length of 96. The results are listed in Table \ref{tab:abl}.


\begin{table}[htbp]
\centering
{\fontsize{9pt}{10pt}\selectfont 
\begin{tabular}{lcc}
\toprule
\textbf{Variants} & \textbf{MSE} & \textbf{MAE} \\
\midrule
w/o ADPT & 0.0220 & 0.1006 \\
w/o EGIA & 0.0211 & 0.0981 \\
w/o KARI & 0.0194 & 0.0941 \\
\textbf{HyperLoad} & \textbf{0.0192} & \textbf{0.0931} \\
\bottomrule
\end{tabular}
}
\caption{Results of the ablation study.}
\label{tab:abl}
\end{table}

\begin{itemize}
\item\textbf{ADPT:} By incorporating text knowledge via cross-modal fusion, the model’s MSE and MAE losses are reduced by 12.7\% and 7.5\%, respectively. This demonstrates that embedding domain-specific background information effectively facilitates the LLMs’ adaptation to data center context, enabling it to leverage pre-training knowledge to encode the complex latent patterns and trends present in multivariate time-series data from data center operations.

\item\textbf{KARI:} Implementing a knowledge-alignment strategy yields additional reductions of 1.0\% and 1.1\% in the model’s MSE and MAE, respectively. This indicates that improving distributional consistency across modalities can effectively reduce cross-modal semantic bias and improves the utilization of textual priors.

\item\textbf{EGIA:} By integrating the EGIA mechanism, the model’s MSE and MAE were reduced by 9.0\% and 5.1\%, respectively, demonstrating that this module can effectively capture the dynamic interrelationships between devices, aid the LLMs in understanding the underlying drivers of load fluctuations, and thereby enhance the accuracy and stability of cooling load forecasting.

\item\textbf{Backbone:} 
In Table \ref{tab:backbones}, we compare the training time and predictive performance of three backbone models (BERT, GPT-2, and LLaMA). LLaMA attains the highest accuracy, with a per-iteration runtime of 0.3928 s, which falls between BERT (0.4289 s) and GPT-2 (0.1943 s), thus effectively balancing speed and precision.



\end{itemize}


\begin{table}[htbp]
  \centering
  {\fontsize{9pt}{10pt}\selectfont 
    \begin{tabular}{cccc}
      \toprule
      Backbones  & Time\,(s / iter) & MSE & MAE \\
      \midrule
      BERT                            & 0.4289           & 0.0217           & 0.1019 \\
      GPT2                  & \textbf{0.1943} & 0.0224           & 0.1012 \\
      \textbf{LLaMA}            & 0.3928           & \textbf{0.0192} & \textbf{0.0931} \\
      \bottomrule
    \end{tabular}%
  }
  \caption{Ablation studies of backbone models.}
  \label{tab:backbones}
\end{table}

\section{Conclusion}

This paper introduces HyperLoad, a framework for forecasting cooling load demand in green data centers. By synergistically combining cross-modal knowledge alignment with multi-scale feature modeling, HyperLoad transfers the long-range dependency modeling and contextual understanding capabilities acquired by LLMs during pre-training to the cooling load prediction task, allowing rapid adaptation to new scenarios and high-accuracy forecasts even under limited data. Experiments show that HyperLoad significantly outperforms mainstream baselines across multiple forecast horizons in both data sufficient and data scarce settings.

\section{Acknowledgments}
This work is supported by the National Natural Science Foundation of China (Grant No.52576234) and the National Key R \& D Program of China (Grant No.2023YFE0108600). This work is also supported by the State Key Laboratory of Alternate Electrical Power System with Renewable Energy Sources (Grant No.LAPS25016).
\bibliography{aaai2026}

\end{document}